\documentclass[sigconf]{acmart}
\acmSubmissionID{738}

\usepackage{booktabs} 

\usepackage{bm}
\usepackage{subcaption}
\usepackage{bbm}
\usepackage{makecell}
\usepackage{multirow}
\usepackage{cuted}

\usepackage[ruled]{algorithm2e} 

\SetAlFnt{\small}
\SetAlCapFnt{\small}
\SetAlCapNameFnt{\small}
\SetAlCapHSkip{0pt}

\copyrightyear{2022}
\acmYear{2022}
\setcopyright{acmlicensed}
\acmConference[SIGGRAPH '22 Conference Proceedings]{Special Interest Group on Computer Graphics and Interactive Techniques Conference Proceedings}{August 7--11, 2022}{Vancouver, BC, Canada}
\acmBooktitle{Special Interest Group on Computer Graphics and Interactive Techniques Conference Proceedings (SIGGRAPH '22 Conference Proceedings), August 7--11, 2022, Vancouver, BC, Canada}
\acmPrice{15.00}
\acmDOI{10.1145/3528233.3530745}
\acmISBN{978-1-4503-9337-9/22/08}

\makeatletter
\def\author@bx@sep{0pc}
\makeatother
\begin{document}
\title{EAMM: One-Shot Emotional Talking Face via Audio-Based Emotion-Aware Motion Model}

\author{Xinya Ji}
\affiliation{%
  \institution{Nanjing University}
  \city{Nanjing}
  \country{China}}
\email{xinya@smail.nju.edu.cn}
\author{Hang Zhou}
\affiliation{%
  \institution{The Chinese University of Hong Kong}
  \city{Hong Kong}
  \country{China}}
\email{zhouhang@link.cuhk.edu.hk}
\author{Kaisiyuan Wang}
\affiliation{%
  \institution{University of Sydney}
  \city{Sydney}
  \country{Australia}}
\email{kaisiyuan.wang@sydney.edu.au}
\author{Qianyi Wu}
\affiliation{%
  \institution{Monash University}
  \city{Melbourne}
  \country{Australia}}
\email{qianyi.wu@monash.edu}
\author{Wayne Wu}
\authornote{Corresponding authors}
\affiliation{%
  \institution{SenseTime Research}
  \city{Shanghai}
  \country{China}}
\email{wuwenyan@sensetime.com}
\author{Feng Xu}
\authornotemark[1]
\affiliation{%
  \institution{BNRist and school of software, Tsinghua University}
  \city{Beijing}
  \country{China}}
\email{xufeng2003@gmail.com}
\author{Xun Cao}
\authornotemark[1]
\affiliation{%
  \institution{Nanjing University}
  \city{Nanjing}
  \country{China}}
\email{caoxun@nju.edu.cn}

\begin{abstract}
Although significant progress has been made to audio-driven talking face generation, existing methods either neglect facial emotion or cannot be applied to arbitrary subjects. 
In this paper, we propose the Emotion-Aware Motion Model (EAMM) to generate one-shot emotional talking faces 
by involving an emotion source video. Specifically, we first propose an Audio2Facial-Dynamics module, which renders talking faces from audio-driven unsupervised zero- and first-order key-points motion. 
Then through exploring the motion model's properties, we further propose an Implicit Emotion Displacement Learner to represent emotion-related facial dynamics as linearly additive displacements to the previously acquired motion representations.
Comprehensive experiments demonstrate that by incorporating the results from both modules, our method can generate satisfactory talking face results on arbitrary subjects with realistic emotion patterns. \footnote{All materials are available at \url{https://jixinya.github.io/projects/EAMM/}.}
\end{abstract}

%
%

\ccsdesc[500]{Computing methodologies~Animation}
\ccsdesc[300]{Computing methodologies~Neural networks}

%
%

\keywords{Facial Animation, Neural Networks}

\begin{teaserfigure}
  \includegraphics[width=\textwidth, scale=0.5]{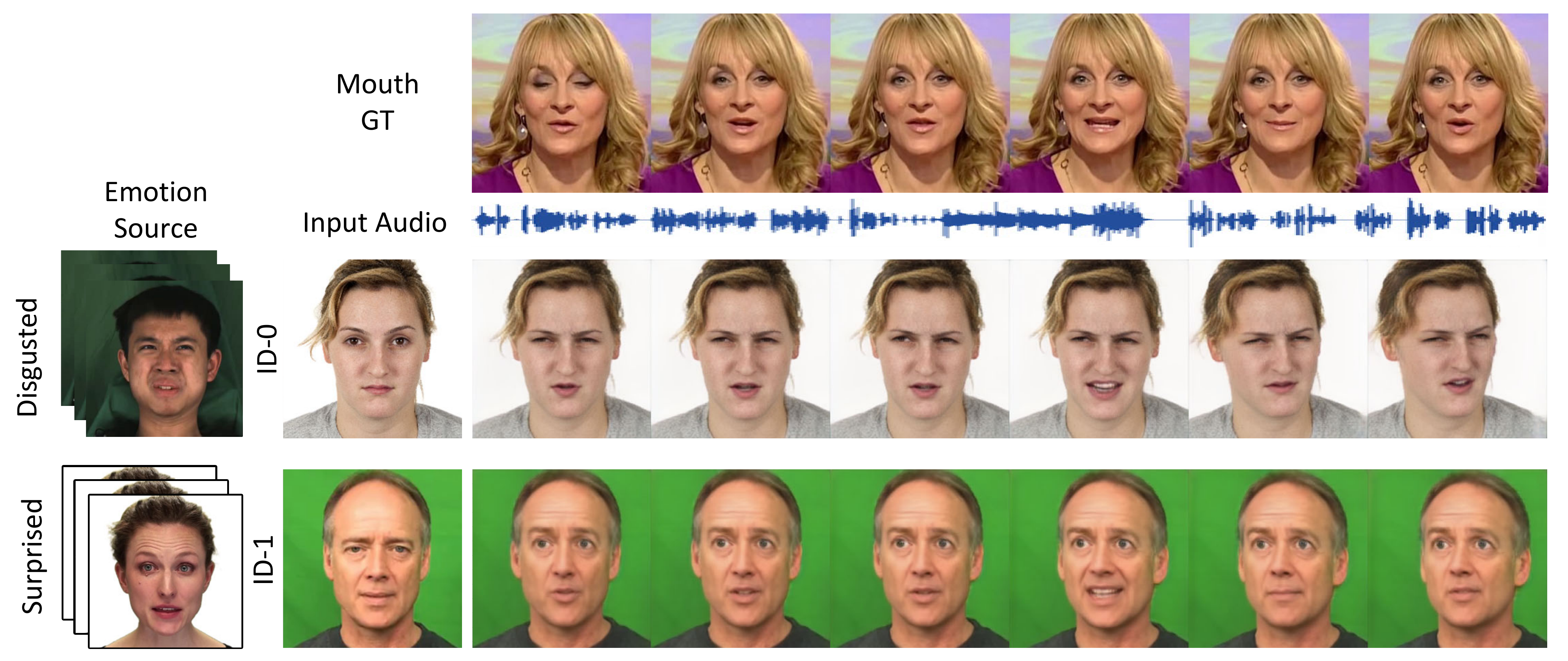}
\captionof{figure}{Qualitative results of our Emotion-Aware Motion Model. Given a single portrait image, we can synthesize emotional talking faces, where mouth movements match the input audio and facial emotion dynamics follow the emotion source video. Natural video (top row) from LRW dataset~\cite{chung2016lip} \copyright \emph{BBC}. Emotional videos (at left corner) from MEAD dataset~\cite{wang2020mead} \copyright \emph{SenseTime} and RAVDESS dataset~\cite{livingstone2018ryerson} \copyright \emph{SMART Lab} (CC BY-NC-SA). Natural face (middle row) from CFD dataset~\cite{ma2015chicago} \copyright \emph{The University of Chicago}. Natural face (bottom row) from CREMA-D dataset~\cite{cao2014crema} (ODbL).}
    \label{fig:first}
\end{teaserfigure}

\maketitle

\section{Introduction}

The task of audio-driven talking face animation enables various applications ranging from visual dubbing, digital avatars, teleconferencing to mixed reality.
While extensive progress has been made in this area~\cite{NVP,fried2019text,zhou2020makelttalk,zhou2021pose}, many of them rely on long video recordings of a source portrait~\cite{edwards2016jali,karras2017audio,yao2021iterative} to generate facial expressions, which are not available in most scenarios. On the other hand, methods driving only one frame \cite{chung2017you,zhou2019talking,mittal2020animating} merely focus on synthesizing audio-synchronized mouth shapes without considering emotion, 
the key factor for realistic animation. 
Thus how to enable expressive emotional editing under the one-shot talking face setting remains an open problem. 

Previous methods either identify emotion from a fixed number of labels~\cite{wang2020mead,abdrashitov2020interactive,li2021write} or only a small range of labeled audio data~\cite{ji2021audio}. However, fixed labels can only represent limited emotions in a coarse-grained discrete manner, making it hard to achieve natural emotion transitions. Additionally, determining emotions from audio only may lead to ambiguities. People sometimes fail to perceive the varying emotions hidden in the speech and the performance of emotion recognition models is not satisfying for general speech. Thus both of them limit the applicability of an emotional talking face model.
Differently, 
we argue that the dynamic emotion can be formulated into a transferable motion pattern extracted from an additional emotional video. 

Therefore, 
our goal is to devise a one-shot talking-face system that takes four kinds of inputs, including an identity source image with neutral expression, a speech source audio, a pre-defined pose and an emotion source video.
However, achieving such a system is not trivial. 1) Generating 
emotional information requires deforming the non-rigid facial structures, which are implicitly but strongly coupled with identity and mouth movements. 
Previous methods usually adopt strong prior of human faces, such as landmarks \cite{kim2019neural,wang2020mead} and 3D models \cite{anderson2013expressive,richard2021audio}. Nevertheless, these methods suffer from error accumulation caused by model inaccuracy.
2) The extraction of emotion patterns is also challenging due to their entanglement with other factors. 

To cope with such issues, in this paper, we present a novel approach named \emph{Emotion-Aware Motion Model (EAMM)}. 
Our intuition is that unsupervised zero- and first-order motion representations~\cite{siarohin2019animating,siarohin2019first,wang2021one}
%
%
are capable of modeling local flow fields on faces, which is suitable for manipulating emotion deformations. 
The key is to \emph{transfer the local emotional deformations to an audio-driven talking face with self-learned key-points and local affine transformations}. Specifically, we firstly achieve talking face generation from a single image through a simple \emph{Audio2Facial-Dynamics (A2FD)} module. It maps audio representations and extracted poses to unsupervised key points and their first-order dynamics. 
%
An additional flow estimator and a generator then cope with the representations for image reconstruction. 

In order to further decompose the local emotion dynamics from appearances, 
we perform empirical explorations 
for the motion model's intrinsic working mechanisms.
Two interesting properties are identified. 1) The dynamic movements in the facial region are only affected by specific key-points and affine transforms, which are denoted as \emph{face-related representations}. 2) The relative displacements of the \emph{face-related representations} are generally linear-additive. 
However, the face-related displacements also contain undesirable mouth movements and structural deformation, making them not directly applicable to our current model.

To this end,
we design an \emph{Implicit Emotion Displacement Learner} to learn only emotion-related displacements on the A2FD module's \emph{face-related representations}.
Particularly, we leverage an effective augmentation strategy on emotion sources to alleviate the influence of undesired factors. 
Then we derive an emotion-feature conditioned implicit function that maps the whole set of motion representations in the A2FD module to the expected \emph{face-related representations}' displacements. 
By linearly combining all motion representations from the two modules, our model complementarily covers both the mouth shapes and emotional dynamics. Extensive experiments demonstrate that our method can generate satisfactory talking face results on arbitrary subjects with realistic emotion patterns.

Our contributions are summarized as follows: \textbf{1)} We propose the Audio2Facial-Dynamics module, which generates neutral audio-driven talking faces by predicting unsupervised motion representations in a simple manner. 
\textbf{2)} Based on two empirical observations, we propose the Implicit Emotion Displacement Learner that can extract the face-related representations' displacements from emotion sources. \textbf{3)} Our proposed Emotion-Aware  Motion  Model (EAMM) manages to generate one-shot talking head animations with emotion control. To the best of our knowledge, it is one of the earliest attempts in this field.

\begin{figure*}[t]
\begin{center}
\includegraphics[width=1.\textwidth]{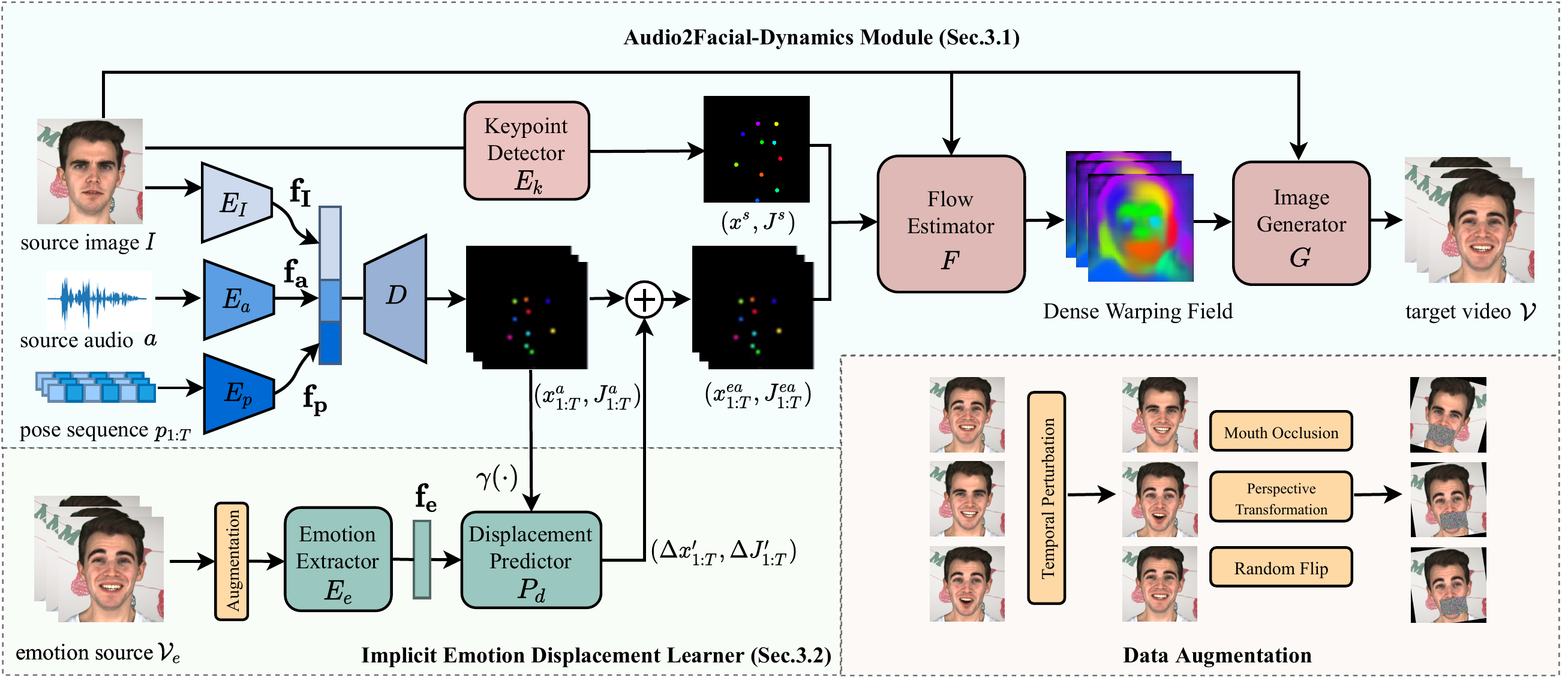}  
\end{center}

\caption{Overview of our \emph{Emotion-Aware Motion Model}. Our framework includes two modules: \emph{Audio2Facial-Dynamics module} for one-shot audio-driven talking head generation and \emph{Implicit Emotion Displacement Learner} for extracting emotional patterns.} 
\label{pipeline}

\end{figure*}

\section{Related Work}
\label{sec:related_work}

\paragraph{Audio-Driven Talking Face Generation.} It is a task which aims to generate talking-face videos from audio clips~\cite{bregler1997video,brand1999voice,wang2012high,VideoRewrite,zhou2018visemenet,lu2021live}. 
These methods can be mainly classified into person-specific or person-agnostic methods. 
Though person-specific methods produce better animation results, their application scenarios are limited. Their training time required for modeling one person may cost several hours~\cite{SIG17Obama} or few minutes~\cite{NVP,lu2021live}. Suwajanakorn et al.~\shortcite{SIG17Obama} synthesize high-quality Obama talking faces from his voice track by using 17-hour videos for training. Thies et al.~\shortcite{NVP} and Lu et al.~\shortcite{lu2021live} propose to generate photo-realistic talking videos with about a 3-minute length person-specific video for training. But they cannot be applied to one image.  On the other hand, Chung et al. \shortcite{chung2017you} generate talking faces in a one-shot manner for the first time. Later, Chen et al. \shortcite{ATVG} and Zhou et al. \shortcite{zhou2020makelttalk} improve the schedule by leveraging facial landmarks as intermediate representations. Zhou et al.~\shortcite{zhou2021pose} further involve pose control into the one-shot setting, but none of these works achieves emotional control.

\paragraph{Emotional Talking Face generation.} Emotion~\cite{cole2017synthesizing} is a factor that plays a strong role in realistic animation. Only a few works consider it in talking face generation due to the difficulty of producing emotion dynamics. Sadoughi et al. \shortcite{sadoughi2019speech} learn the relationship between emotion and lip movements from a designed conditional generative adversarial network. Vougioukas et al. \shortcite{vougioukas2020realistic} introduce three discriminators for a temporal GAN. However, both of them fail to generate semantic expressions and achieve emotion manipulation. Recently, Wang et al. \shortcite{wang2020mead} collect the MEAD dataset and set emotions as one-hot vectors to achieve emotion control. While Ji et al. \shortcite{ji2021audio} propose to decompose speech into decoupled content and emotion spaces, and then synthesize emotion dynamics from audio. Nevertheless, their methods cannot be applied to unseen characters and audios. Different from them, we resort to a source video and disentangle the emotion information to achieve emotion control in the one-shot setting.

\paragraph{Video-driven Facial Animation.}
Video-driven animation leverages a video to reenact facial motion, which is highly related to audio-driven talking-face generation. Traditional approaches demand prior knowledge or manual labels of the animated target such as 3D morphable model~\cite{thies2016face2face,kim2018deep,zollhofer2018state} or 2D landmark~\cite{isola2017image,wu2018reenactgan,huang2020learning,zhang2020freenet,burkov2020neural,chen2020puppeteergan,zakharov2020fast,yao2020mesh,tripathy2021facegan}. Recently, a few methods \cite{siarohin2019animating,siarohin2019first} that do not require priors have been explored. They employ a self-supervised framework and model the motion in a dense field, in which case appearance and motion are decoupled. Our model is built upon a similar idea.


\section{Method}

The overview of our Emotion-Aware Motion Model (EAMM) is shown in Figure~\ref{pipeline}, where different kinds of signals are taken as the inputs to generate emotional talking faces. Our EAMM mainly consists of two parts, an \emph{Audio2Facial-Dynamics module} that achieves audio-driven talking face generation with neutral expressions from one neutral frame (Section~\ref{sec:3.1}) and an \emph{Implicit Emotion Displacement Learner} that involves emotional dynamics (Section~\ref{sec:3.2}).
In the following sections, we introduce each part in detail.

\subsection{Audio2Facial-Dynamics Module}
\label{sec:3.1}
The first step toward audio-driven emotional talking face is to build a one-shot system that is plausible for integrating expression dynamics. To this end, we design the Audio2Facial-Dynamics (A2FD) module, which firstly
models facial movements with neutral expressions. The motion is represented as a set of unsupervised key-points and their first order dynamics inspired by \cite{siarohin2019first,wang2021audio2head}. Based on this motion representation, warping fields can be calculated to account for the local facial motions, and thus facilities further the generation of emotional talking faces.

\paragraph{Training Formulation.}
Since direct supervision is not available due to the lack of paired data, we adopt the self-supervised training strategy~\cite{chen2019hierarchical,zhou2021pose}. For each training video clip $\mathcal{V} = \{\bm{I}_{1},...\bm{I}_{t},...\bm{I}_{T}\}$, we randomly select one frame $\bm{I}$ as the identity source image, and take Mel Frequency Cepstral Coefficients (MFCC)~\cite{logan2000mel} $\bm{s}_{1:T}$  of the corresponding speech audio $\bm{a}$ as the speech source audio representations.
Considering that head pose is also a key component, which can hardly be inferred from audios, we assign the pose sequence $\bm{p}_{1:T}$ estimated from the training video clip by an off-the-shelf tool \cite{guo2020towards} as additional inputs. A 6-dim vector (i.e., 3 for rotation, 2 for translation and 1 for scale) is used to represent head pose for each frame $\bm{p}_{t}$. 
Note that in the testing stage, the identity image $\bm{I}$, the speech source audio clip $\bm{a}$ and the pose sequence $\bm{p}_{1:T}$ can come from \emph{different sources}.

\paragraph{Pipeline of A2FD} 
As illustrated in Figure~\ref{pipeline}, we first use three encoders (i.e., $\bm{E}_{I}$, $\bm{E}_{a}$ and $\bm{E}_{p}$) to extract the corresponding information from the three inputs, which are denoted as the identity feature $\mathbf{f}_{I}$, the audio feature $\mathbf{f}_{a}$ and the pose feature $\mathbf{f}_{p}$. Then we combine the three extracted features and feed them into a LSTM-based~\cite{hochreiter1997long} decoder $\bm{D}$ to recurrently predict the unsupervised motion representations for the whole sequence. 
The motion representations at each time step $t$ are composed of $N$ implicitly learned key-points $\bm{x}^{a}_{t} \in \mathbb{R}^{N\times2}$ and their first order motion dynamics, i.e., jacobians $\bm{J}^{a}_{t} \in \mathbb{R}^{N\times2\times2}$, where each jacobian denotes the local affine transformation for the neighbourhood area at each key-point (zero-order representation) position. We set $N = 10$ by default throughout the paper. 

In order to derive the warping fields correlated with local dynamics, the standard-positioned zero- and first-order representations of the initial frame $\bm{I}$ should be provided. Instead of learning all representations from scratch, we argue that our A2FD module would be easier to learn if we share the audio-involved key-point distribution with a pretrained video-driven first-order motion model's~\cite{siarohin2019first}.

Thus we employ a pretrained key-point detector $\bm{E}_{k}$ from \cite{siarohin2019first} to predict the initial motion representations $\bm{x}^{s}$ and $\bm{J}^{s}$ from the source image $\bm{I}$. 
%
Then we adopt a flow estimator $\bm{F}$ to generate a dense warping field that describes the non-linear transformation from the source image to the target video frame. Specifically, at each time step $t$, we first calculate $N$ warping flows as well as a set of masks $\mathbf{M}$ based on the predicted key-points $\bm{x}^{a}_{t}$, $\bm{x}^{s}$ and the jacobians $\bm{J}^{a}_{t}$, $\bm{J}^{s}$. Then by weighted combining the masks $\mathbf{M}$ to the warping flows, we obtain the final dense warping field. Finally, we feed the dense warping field together with the source image $\bm{I}$ into an image generator $\bm{G}$ to produce the final output frame at each time step $\bm{\hat{I}}_t$. Please refer to \cite{siarohin2019first} for more details.

\paragraph{Training Objectives.}
As stated before, we would like to share the motion representation's distribution with the visual-based model, we leverage $\bm{E}_{k}$ as a specific teacher network for our audio-based model learning. Specifically, the key-points $\bm{x}^{v}_{t}$, and their jacobian $\bm{J}^{v}_{t}$ extracted by $\bm{E}_{k}$ from the training video clip $\mathcal{V}$ are served as intermediate supervisions. 
Then we formulate a key-point loss term $L_{kp}$ defined below to train our A2FD module:
\begin{equation}
\label{kp}
\begin{aligned}
L_{kp} &= \frac{1}{T}\sum_{t=1}^{T}(\| \bm{x}^{a}_{t} - \bm{x}^{v}_{t} \|_1 +\| \bm{J}^{a}_{t} - \bm{J}^{v}_{t} \|_1).
\end{aligned}
\end{equation}
In the second stage, we use a perceptual loss term $L_{per}$ to fine-tune the model by minimizing the difference between the reconstructed frame $\bm{\hat{I}_t}$ and the target frame $\bm{I}_t$:
\begin{equation}
\label{per}
\begin{aligned}
L_{per}& = \sum_{i=1}^{l}\| \text{VGG}_{i}(\bm{\hat{I}}_t - \text{VGG}_{i}(\bm{I}_t))\|_1, \\
\end{aligned}
\end{equation}
where $\text{VGG}_{i}(\cdot)$ is the $i^{th}$ channel feature of a pretrained VGG network \cite{johnson2016perceptual} with $l$ channels. The total loss function is defined as:
\begin{equation}
\begin{aligned}
L_{mo} &= L_{kp} + \lambda_{per}L_{per}, \\
\end{aligned}
\end{equation}
where $\lambda_{per}$ represents the weight for $L_{per}$. 
%


\begin{figure}[t]
\begin{center}
\includegraphics[width=.95\linewidth]{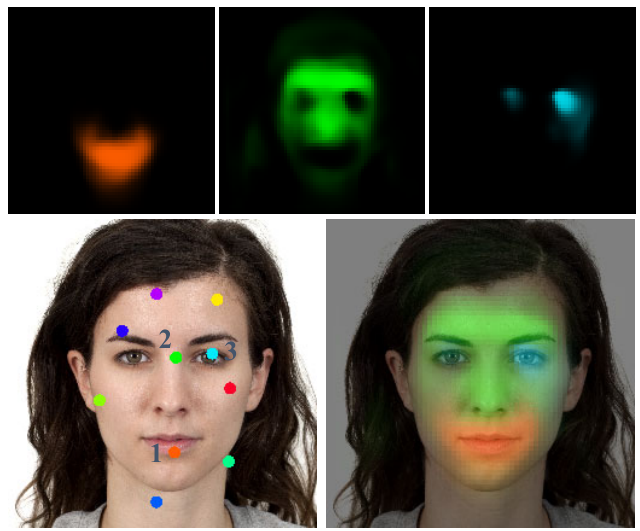} 
\end{center}

\caption{Visualization of the masks for face related key-points. The bottom-left image shows ten learned key-points and the bottom-right image indicates the composition masks. We also visualize the masks of the three face-related key-points separately in the top row. Natural face from CFD dataset \copyright \emph{The University of Chicago}.}

\label{disentangle}
\end{figure}

\begin{table*}
\caption{\small{Quantitative comparisons with state-of-the-art methods. We show quantitative results on LRW \cite{chung2016lip} and MEAD \cite{wang2020mead} datasets. The results of LRW are only generated by the Audio2Facial-Dynamics Module, as there is no emotion annotation in LRW. The metrics related to video quality and landmarks are calculated by comparing the generated results with the ground truth. Here M- denotes mouth and F- denotes face region. The signages ``$\uparrow$'' and ``$\downarrow$'' indicate higher and lower metric values for better results, respectively.} }

\label{metric}
\begin{center}
\begin{tabular}{l| c c c c c |c c c c c}
\toprule
\multirow{2}*{Method} & \multicolumn{5}{c}{LRW \shortcite{chung2016lip}} & \multicolumn{5}{c}{MEAD \shortcite{wang2020mead}} \\
\cmidrule(lr){2-6} \cmidrule(lr){7-11}
& SSIM $\uparrow$ & PSNR $\uparrow$ & SyncNet $\uparrow$ & M-LMD $\downarrow$ & F-LMD $\downarrow$ & SSIM $\uparrow$ & PSNR $\uparrow$ & SyncNet $\uparrow$ & M-LMD $\downarrow$ & F-LMD $\downarrow$\\
\hline 
ATVG \shortcite{chen2019hierarchical}   &0.69   &28.98   &5.12    &2.69 &2.64   &0.57   &28.58   &2.24    &3.14 &3.87\\
SDA \shortcite{vougioukas2018end}      &0.50   &28.96   &5.26    &2.75 &4.74   &0.44   &28.54   &1.88    &3.99 &4.50\\
Wav2Lip \shortcite{prajwal2020lip}      &0.73   &30.63   &\textbf{5.94}    &1.81 &2.46   &0.57   &29.03   &2.24    &3.43 &3.80\\
MakeItTalk \shortcite{zhou2020makelttalk}   &0.69   &30.38  &5.34 &2.20  &2.83  &0.56   &28.92   &2.20 &3.80 &3.92\\
PC-AVS \shortcite{zhou2021pose}        &0.71   &30.39   &5.46    &1.67 &\textbf{2.05}  &0.60   &29.02   &2.10    &2.97 &2.74\\
Ground Truth     &1.00   &-  &5.56    &0.00  &0.00   &1.00   &- &2.18  &0.00  &0.00\\
Ours             &\textbf{0.74}   &\textbf{30.92}   &5.52  &\textbf{1.61}  &2.08   &\textbf{0.66}   &\textbf{29.29} &\textbf{2.26}   &\textbf{2.41}   &\textbf{2.55}\\
\bottomrule
\end{tabular}

\end{center}
\end{table*}

\paragraph{Discussion.} After the generation of neutral talking faces with audio inputs, one straightforward idea is to directly incorporate the emotional source into this pipeline. However, an emotional source naturally contains all facial information including the mouth, identity and pose, leading to undesirable results. Thus this brings the need to decouple emotional information within our motion representations and warping field.

We start by exploring how the warping field transforms the source image $\bm{I}$ based on the key-points $\bm{x}$.
We visualize the composition masks $\mathbf{M}$ shown in Figure~\ref{disentangle} and observe that the face region is only affected by three \emph{face-related} key-points. The set of representations with only three key-points are denoted as $(\bm{x}', \bm{J}')$.

Inspired by this observation, we perform a simple experiment to validate whether we can transfer the emotion pattern from an emotion source video to our A2FD module by merely editing the three face-related key-points and their jacobians.
One simple idea is to find whether the deviation between emotional and neutral motion representations of the same person can be linearly additive, \textit{i.e.}, to impose emotion by adding the displacements on other faces' motion representations.
To alleviate the influence of the mouth, we leverage both the pretrained model that extracts full facial dynamics and our A2FD model that generates neutral talking faces. Ideally, their mouth shapes should be aligned within the representations.

Concretely, we first detect key-points ${\bm{x}^{e}}'$ and jacobians ${\bm{J}^{e}}'$ from an emotion source video with $\bm{E}_{k}$. 
Then we feed its audio and a 
neutral status image of this person
into our A2FD module to generate ${\bm{x}^{n}}'$ and ${\bm{J}^{n}}'$. 
We calculate the deviation (${\bm{x}^{e}}' - {\bm{x}^{n}}'$, ${\bm{J}^{e}}' - {\bm{J}^{n}}'$), which is assumed to include emotion information. 
By simply adding this deviation as displacements onto the motion representations of an arbitrary person, we observe that the motion dynamics can be successfully transferred on the generated results. Thus we can regard these representations as roughly linearly additive.

However, while the emotion information can be preserved, we observe that there are many undesirable artifacts around the face boundary and the mouth. 
A possible explanation is the calculated displacements include not only emotion information but also other factors, such as identity, pose and speech content, which results in inaccurate guidance for the subsequent generation.

\subsection{Implicit Emotion Displacement Learner}
\label{sec:3.2}
According to the observation above,
we can basically formulate the emotion pattern as the complementary displacements to the face-related key-points and jacobians.
Therefore, we design an Implicit Emotion Displacement Learner to extract emotion information from the emotional video $\mathcal{V}_{e} = \{\bm{Q}_{1},...\bm{Q}_{t},...\bm{Q}_{T}\}$ and then encode them as the displacements $(\Delta \bm{x}', \Delta\bm{J}')$ to the three face-related key-points and jacobians $(\bm{x}', \bm{J}')$ from the A2FD module.

\paragraph{Data Processing.}
To disentangle the emotion from other factors, we design a special data augmentation strategy. Specifically, to block the speech content information, we occlude the lip and jaw movements using a mask filled with random noise. In addition, to eliminate the effects of pose and natural movement like blinking, we introduce a \emph{temporal perturbation technique}. For each time step $t$, instead of using the frame $\bm{Q}_{t}$ for emotion extraction, we select a frame from different time steps perturbed around the current time $t$. Moreover, to further alleviate the influence of facial structure information, we apply the perspective transformation and random horizontal flip~\cite{zhou2021pose}. 
This data augmentation strategy is also demonstrated in Figure~\ref{pipeline}.

\paragraph{Learning Emotion Displacements.}
To incorporate the emotion pattern into our A2FD module, we first employ an emotion extractor $\bm{E}_{e}$ to extract the emotion feature $\mathbf{f}_{e}$ from the processed video frames. For generating emotion dynamics synchronized with the input audio, 
we take the key-points $\bm{x}^{a}_{1:T}$ and their jacobians $\bm{J}^{a}_{1:T}$ predicted from the A2FD module together with $\mathbf{f_{e}}$ as the inputs to our displacement predictor $\bm{P}_{d}$. It employs a 4-layer multiple layer perceptrons (MLP) to predict the displacements referred to as $\Delta \bm{x}^{a'}_{1:T}$ and $\Delta \bm{J}^{a'}_{1:T}$.
Note that a positional encoding operation \cite{mildenhall2020nerf} is performed to project the key-points into a high dimensional space, which enables the model to capture higher frequency details.
Finally, we produce the $N$ emotional audio-learned key-points $\bm{x}^{ea}_{1:T}$ and jacobians $\bm{J}^{ea}_{1:T}$ by linearly adding $\Delta \bm{x}^{a'}_{1:T}$ and $\Delta \bm{J}^{a'}_{1:T}$ onto audio-learned representations $\bm{x}^{a'}_{1:T}$, $\bm{J}^{a'}_{1:T}$.

\begin{figure*}[t]
\begin{center}
\setlength{\abovecaptionskip}{0.cm}
\includegraphics[width=1.\textwidth]{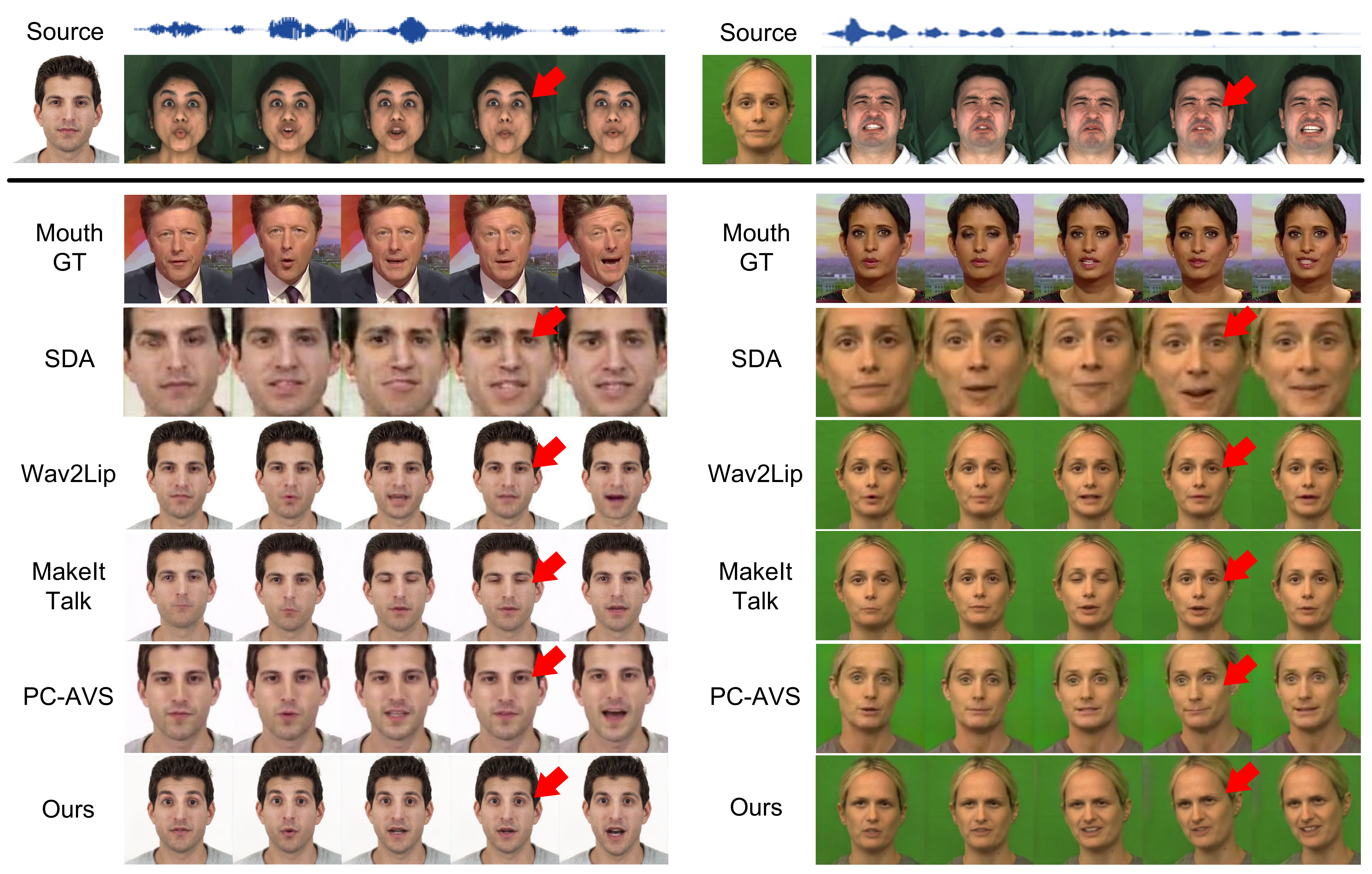} 
\end{center}

\caption{Qualitative comparisons with state-of-the-art methods on two cases. The top row shows the identity, speech content audio, and emotion source clips. The emotion categories of the videos are surprised (left) and angry (right). The second row shows the corresponding frames of the audio source, \textit{i.e.}, ground truth for mouth shapes. Please zoom in to see the expressions at the red arrows. Natural videos (second row) from LRW dataset \copyright \emph{BBC}. Natural face (left) from CFD dataset \copyright \emph{The University of Chicago}. Natural face (right) from CREMA-D dataset (ODbL).} 
\label{comparison}

\end{figure*}

\paragraph{Training Objectives.}
During training, we follow the self-supervised training strategy in Sec.~\ref{sec:3.1}.
Specifically, for each emotion source video $\mathcal{V}_{e}$, we use the pretrained detector $\bm{E}_{k}$ to extract the $N$ key-points $\bm{x}^{e}_{1:T}$ and jacobians $\bm{J}^{e}_{1:T}$ as the ground-truth, 
and then we minimize the difference between the emotional audio-learned key-points ${\bm{x}}^{ea}_{1:T}$, jacobians $\bm{J}^{ea}_{1:T}$ and the ground-truth by reformulating the loss term $L_{kp}$ in Eq.~\ref{kp} as:
\begin{equation}
\label{kp_pro}
\begin{aligned}
L_{kp} &= \frac{1}{T}\sum_{t=1}^{T}(\| \bm{x}^{ea}_{t} - \bm{x}^{e}_{t} \|_1 +\| \bm{J}^{ea}_{t} - \bm{J}^{e}_{t} \|_1).
\end{aligned}
\end{equation}
Note that we also use the loss $L_{per}$ in Eq.~\ref{per} to finetune the A2FD module when training our implicit emotion displacement learner.

\subsection{Implementation Details}
All videos are aligned via centering (crop \& resize) the location of the first frame's face and resized to 256 $\times$ 256. The videos are sampled at the rate of 25 FPS and the audios are pre-processed to 16kHz. For audio features, we compute 28-dim MFCC with the window size of 10 ms to produce a 28 $\times$ 12 feature for each frame.

\paragraph{Dataset.}
We use the LRW \cite{chung2016lip} dataset which has no emotion annotation to train our A2FD module. LRW is an in-the-wild audio-visual dataset collected from BBC news, including 1000 utterances of 500 different words, each of which lasts for about 1 second. Featuring a variety of speakers and head motions, it is well-suited for our training objectives. We split the train/test corpus following the setting of LRW.

Emotional dataset MEAD~\cite{wang2020mead} is used to train our implicit emotion displacement learner. MEAD is a high-quality emotional talking-face dataset, including recorded videos of different actors speaking with 8 different emotions. Here, we select 34 actors for training and 6 actors for testing.



\begin{table*}
\caption{ User study results evaluated on three different aspects (the maximum score value is 5) and the emotion classification accuracy.}

\label{user}
\begin{center}
\begin{tabular}{l|c| c| c| c| c |c| c}
\toprule
Score/Method & ATVG \shortcite{chen2019hierarchical} & SDA \shortcite{vougioukas2018end} & Wav2Lip \shortcite{prajwal2020lip} & MakeItTalk \shortcite{zhou2020makelttalk} & PC-AVS \shortcite{zhou2021pose} & Real & Ours\\
\hline 
Lip Synchronization   &3.01   &2.30   &3.54   &3.48 &3.79   & 4.41 &\textbf{3.81}   \\
Facial Expression Naturalness   &2.52  &2.25   &2.65  &3.34  &3.10  & 4.03 &\textbf{3.47}\\
Video Quality      &2.54   &2.17   &3.83  &3.79  &3.26   & 4.42 &\textbf{3.89}  \\
Emotion Accuracy      &10\%   &11\%   &13\%  &12\%  &15\%   &71\%   &\textbf{58\%}  \\
\bottomrule
\end{tabular}

\end{center}
\end{table*}

\section{Results}

In the following, we present the comparison results with other state-of-the-art methods, the results of a user study and the design evaluation of our approach. Please see the supplementary for more details of the experiment settings.

\subsection{Evaluation}
\label{sec:4.1}
We perform comparisons with state-of-the-art methods (i.e., ATVG \cite{chen2019hierarchical}, Speech-driven-animation \cite{vougioukas2018end}, Wav2Lip \cite{prajwal2020lip},  MakeItTalk \cite{zhou2020makelttalk}, PC-AVS \cite{zhou2021pose}) on the test set of LRW and MEAD.

\paragraph{Evaluation Metrics.}
To evaluate the synchronization between the generated mouth shapes and the input audio, we adopt the metric landmarks distances on the mouth (\emph{M-LMD}) \cite{chen2019hierarchical} and the \emph{confidence score of SyncNet} \cite{Chung16a}. Then we use the LMD on the whole face (\emph{F-LMD}) to measure the accuracy of facial expression and poses. To evaluate the quality of the generated videos, we also introduce \emph{SSIM} \cite{wang2004image} and \emph{PSNR} as additional metrics.

\paragraph{Quantitative Results.}
The experiments are conducted in a self-driving setting, in which we use the audio and detected pose sequence of each test video as the audio and pose source.
Note that for the LRW dataset without emotion, we only use the A2FD Module to generate the results, 
where we randomly select a frame from each video in LRW as the source image.
While for the MEAD dataset with emotions, the source image is randomly selected from a neutral video of the same speaker as in the test video. 
Moreover, instead of directly using the test video as the emotion source, we adopt a fair setting as in~\cite{zhou2021pose} for emotion source acquisition.
We first align all the generated and real frames to the same size and then detect their facial landmarks for comparison.
The comparison results are reported in Table~\ref{metric}. Our method achieves the highest score among all metrics on MEAD and most metrics on LRW. 
It's worth noting that Wav2Lip is trained with a SyncNet discriminator, thus it is natural to get the highest confidence score of SyncNet on LRW. Our results are comparable with the ground truth, which means the achievement of satisfactory audio-visual synchronization.
As for the F-LMD that accounts for both pose and expressions, our method achieves comparable results with PC-AVS on LRW. The reason is that there are fewer emotional expression changes on LRW compared with MEAD, on which we achieve better results.

\paragraph{Qualitative Results.}
We also provide a qualitative comparison between our method and state-of-the-art methods in Figure~\ref{comparison}. 
Here we randomly select an emotional video in MEAD as the emotion source for our method. Our method can generate vivid emotional animation with natural head movements and accurate mouth shapes, while other methods cannot generate obvious emotional dynamics (see the red arrows). Concretely, only Wav2Lip and PC-AVS can generate mouth motions competitive with ours. However, Wav2Lip merely focuses on the synchronization between the speech audio and lip movements without considering the facial expression and head pose. Though PC-AVS is able to control the head poses, it neglects emotion dynamics for generating realistic animation. SDA can produce results with changing facial expressions, however, the generated expressions are always unstable which affects the identity.


\subsection{User Study}
\label{sec:4.2}

We conduct a user study to compare our method with real data and other state-of-the-art methods mentioned before. We recruit 20 participants with computer science background, in which 14 are male and 6 are female. The ages of the participants range from 21 to 25. We select 5 videos for each emotion category on the test set of MEAD as the emotion source videos. For each emotion source video, we randomly select the image and audio source from the test set of LRW and MEAD to generate 40 videos (5 $\times$ 8 emotions) for each method. We also randomly select 40 real videos with corresponding emotions. Thus, each participant engages in 280 (7 $\times$ 40 videos) trials and the videos are shown in a random order to reduce fatigue. We first show real labeled videos with eight different emotion categories to the participants for reference. Then, for each presented video clip, we design a two-stage procedure. In the first stage, the participants are required to evaluate the given video from three aspects (i.e., ``lip synchronization'', ``naturalness of facial expression'', and ``video quality'') and give a score from 1 (worst) to 5 (best) for each aspect.

Moreover, since there are specific emotion labels for source videos in MEAD, we conduct an emotion classification task to evaluate the generated emotion on our method in the second stage. Specifically, we show the same muted video and ask the participants to choose the emotion type of the video from eight categories. The videos shown in the second stage are muted so that the participants can only focus on the facial expressions. The generated video and emotion can be well evaluated in this way. Basically, it takes about 90 minutes for each participant to finish the experiment.

The results are shown in Table~\ref{user}.
Our work achieves the highest scores over the three aspects apart from the real data, which indicates the effectiveness of our method. 
What's more, we obtain 58\% accuracy in emotion classification, while the accuracy scores of other methods are much lower than ours, since they cannot generate realistic emotion dynamics.


\subsection{Ablation Study}
\label{sec:4.3}

We conduct ablation studies on the MEAD dataset to demonstrate the effectiveness of our \textit{Implicit Emotion Displacement Learner} (Section~\ref{sec:3.2}) and verify the contributions of the three important components in it (i.e., data augmentation, conditional audio-driven key-points and jacobians input and learning displacements for \textit{three face-related representations}). Specifically, we design five variants in total, and the first two of them are designed to evaluate our motion model design: (1) \textit{A2FD (Baseline) }: our EAMM without using \textit{Implicit Emotion Displacement Learner}; (2) \textit{Feature-based}: representing emotion dynamics at the feature space. The other three are designed to verify the components in \textit{Implicit Emotion Displacement Learner}: (3) \textit{w/o augmentation}: without using data augmentation; (4) \textit{w/o condition}: without using the conditional audio-driven key-points and jacobians input; (5) \textit{displacement for all points}: learning emotion displacements for all the key-points and their jacobians.
Note that the variant \textit{Feature-based} is designed to explore whether the emotion pattern can be represented as a feature instead of the displacement manner. Specifically, we first use two separated encoders to extract the audio feature $\mathbf{f}_{a}$ and the emotion feature $\mathbf{f}_{e}$, respectively. Then we introduce a commonly used operation in the works of style-transfer, AdaIN \cite{huang2017arbitrary}, to transfer the emotion style from the emotion feature $\mathbf{f}_{e}$ to the audio feature $\mathbf{f}_{a}$. Lastly, we use a decoder \cite{siarohin2019first} to predict the final key-points and jacobians.

\begin{figure}[t]
\begin{center}
\includegraphics[width=0.95\linewidth]{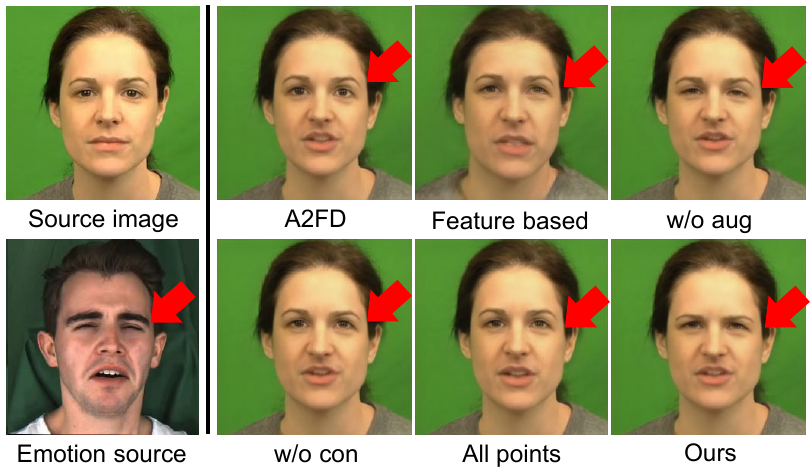}

\end{center}
\caption{Qualitative results of ablation study. We use the same source image for all the variants and the pose is set to be static for better visualization. Natural face from CREMA-D dataset (ODbL).}

\label{ablation}

\end{figure}

Apart from the metrics mentioned in Section~\ref{sec:4.1}, we additionally use an off-the-shelf emotion classification network \cite{meng2019frame} to evaluate the accuracy of the generated emotion. The classification network is trained on the MEAD dataset and achieves a 90\% accuracy rate on the test set, which ensures the confidentiality of the evaluation results. The quantitative results shown in Table~\ref{Ablation_study} and the visualization shown in Figure~\ref{ablation}, both demonstrate that \textit{Implicit Emotion Displacement Learner} and its three components are effective designs for emotion generation. Among the three components, the data augmentation strategy is especially critical for our model, as it contributes to transferring accurate emotional dynamics without sacrificing the identity (see the red arrows). Moreover, we observe that the face shape produced by the feature-based model is unstable and the emotion is not obvious either, which indicates that emotion cannot be well disentangled at the feature space. 


\begin{table}[t] 

\caption{Quantitative ablation study. We provide quantitative results of five variants and our EAMM on emotion accuracy, M-LMD, F-LMD and SSIM. }

\label{Ablation_study}
\begin{center}
\begin{tabular}{l|c|c|c|c} 
\toprule
Method/Score &Acc$_{emo}$ $\uparrow$ & M-LMD$\downarrow$  & F-LMD$\downarrow$  & SSIM $\uparrow$ \\
\hline 
A2FD (Baseline)          &14.07    &2.78   &2.69   &0.62\\
Feature-based     &48.26    &2.62   &2.61   &0.64   \\ 
\hline
w/o augmentation  &49.37    &2.60   &2.56   &0.63\\
w/o condition     &52.54    &2.58   &2.60   &0.65\\
All points        &57.26    &2.49   &2.62   &\textbf{0.66}\\
Ours           &\textbf{68.41}      &\textbf{2.41}   &\textbf{2.55}  &\textbf{0.66}\\
\bottomrule
\end{tabular}

\end{center}
\end{table}

\subsection{Limitations} Despite the success of our approach, we also recognize some limitations during the exploration.
Firstly, the emotion dynamics generated in the mouth region are not obvious in our work due to the mouth occlusion operation in our data augmentation strategy.
Secondly, since emotion is a personalized factor, the emotion pattern extracted from a character sometimes seems to be unnatural after being transferred to another one. 
Moreover, our method neglects the correlation between the audio and emotion by involving an emotion source video which may lead to incongruent animation results.
These will be part of our future work.

\section{Conclusion}

In this paper, we present the Emotion-Aware Motion Model to generate one-shot emotional talking faces by incorporating an additional emotion source video. 
We extract the emotion dynamics which is formulated as a transferable motion pattern in the emotion source video and apply it to arbitrary audio-driven talking face.
This allows us to synthesize more realistic talking faces which has great potential in applications, e.g., video conferencing, digital avatars.
Qualitative and quantitative experiments show that our method can generate more expressive animation results compared with state-of-the-art methods.
We hope our work can inspire future research in this field.

\section{Ethical Considerations}
Our method focuses on synthesizing emotional talking face animation, which is intended for developing digital entertainment and an advanced teleconferencing system.
However, it may also be misused for some malicious proposes on social media, which leads to negative impacts on the whole society.
To alleviate the concerns above, significant progress has been made in the Deepfake detection area.
%
Some works~\cite{guera2018deepfake, rossler2019faceforensics++, li2020face, yu2019attributing, wang2020cnn, chai2020makes} focus on identifying visual deepfakes by detecting texture artifacts or inconsistency. Recent studies~\cite{owens2016ambient,arandjelovic2017look,korbar2018cooperative, zhou2021joint} also consider the relationship between the video and audio and use the synchronization of these two modalities to benefit the detection.
However, the lack of large realistic and emotional portrait data limits their performance and generalizability.
Therefore, we are also committed to supporting the Deepfake detection community by sharing our generated emotional talking face results, which can improve the detection algorithms to handle more complex scenarios.
We believe that the proper use of this technique will enhance positive societal development in both machine learning research and daily life.

\begin{acks} 
  This work was supported by the NSFC (No.62025108, 62021002, 61727808),  the NSFJS (BK20192003), the Beijing Natural Science Foundation (JQ19015), the National Key R\&D Program of China 2018YFA0704000. This work was supported by the Institute for Brain and Cognitive Science, Tsinghua University (THUIBCS) and Beijing Laboratory of Brain and Cognitive Intelligence, Beijing Municipal Education Commission (BLBCI). 
\end{acks}

\bibliographystyle{ACM-Reference-Format}
\bibliography{sample-bibliography}


\end{document}